\documentclass{article}
\usepackage{spconf,amsmath,graphicx,diagbox}
\usepackage{caption}
\usepackage{subcaption}

\title{Intent Recognition and Unsupervised Slot Identification \\for low-resourced Spoken Dialog Systems}
\name{\textit{\textit{A. Gupta, O. Deng$*$, A. Kushwaha$*$, S. Mittal$*$ W. Zeng$*$, S.k. Rallabandi, A.W. Black}}\thanks{$*$Equal Contribution.}}
\address{\textit{\{akshatgu, odeng, akrutik, salonim, wwzeng, srallaba, awb\}@andrew.cmu.edu}\\Carnegie Mellon University}
\begin{document}

\maketitle

\begin{abstract}
Intent Recognition and Slot Identification are crucial components in spoken language understanding (SLU) systems. In this paper, we present a novel approach towards both these tasks in the context of low-resourced and unwritten languages. We use an acoustic based SLU system that converts speech to its phonetic transcription using a universal phone recognition system. We build a word-free natural language understanding module that does intent recognition and slot identification from these phonetic transcription. Our proposed SLU system performs competitively for resource rich scenarios and significantly outperforms existing approaches as the amount of available data reduces. We train both recurrent and transformer based neural networks and test our system on five natural speech datasets in five different languages. We observe more than 10\% improvement for intent classification in Tamil and more than 5\% improvement for intent classification in Sinhala. Additionally, we present a novel approach towards unsupervised slot identification using normalized attention scores. This approach can be used for unsupervised slot labelling, data augmentation and to generate data for a new slot in a one-shot way with only one speech recording. 
\end{abstract}
\section{Introduction}
\label{sec:intro}

\begin{figure}
     \centering
     \begin{subfigure}{\linewidth}
         \centering
         \includegraphics[width=\linewidth]{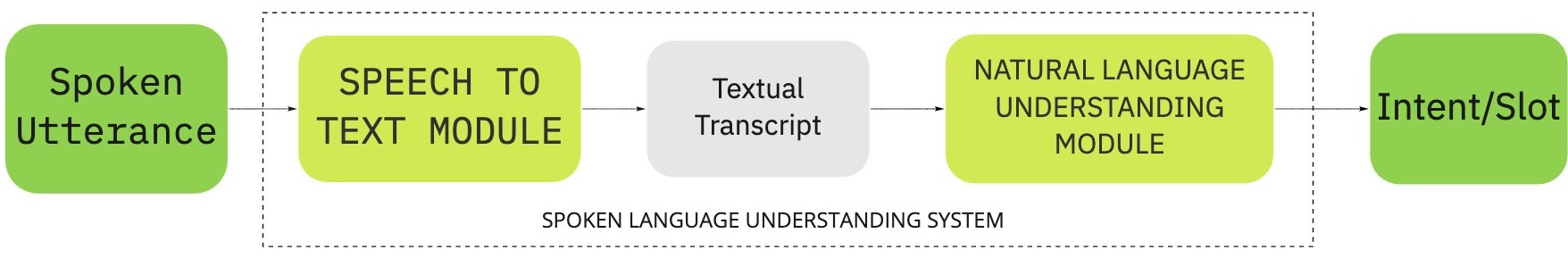}
         \caption{Block diagram of a typical spoken language understanding system}
         \label{fig:SLU}
     \end{subfigure}
     \hfill
     \begin{subfigure}{\linewidth}
         \centering
         \includegraphics[width=\linewidth]{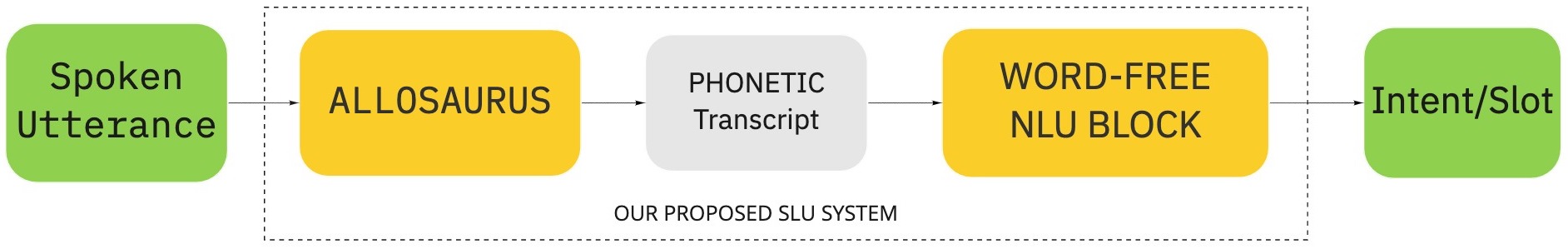}
         \caption{Block diagram representing our proposed SLU system.}
         \label{fig:NOW}
     \end{subfigure}
        \caption{Diagrammatic description of a typical SLU system and our proposed SLU system.}
\end{figure}

Spoken dialog systems are slowly integrating themselves in everyday human lives, being used for various applications that include accessing information, doing transactions, tutoring and entertainment. Speech is presumably the most natural form of interaction for humans. Spoken dialog systems not only create a very natural interface for humans to interact with technology, but also overcome the barriers posed by a written interface. Thus, access to technology is not restricted by literacy and can also be done in unwritten languages. Currently, spoken dialog systems are available only in a limited number of languages. A major bottleneck in extending these systems to other low-resourced, local and unwritten languages is the lack of availability annotated data in these languages. 

Spoken language understanding (SLU) systems are fundamental building blocks of spoken dialog systems. A typical SLU pipeline, shown in Figure \ref{fig:SLU}, comprises of a Speech-to-Text (STT) module followed by a Natural Language Understanding (NLU) module. STT modules convert speech to textual transcriptions and NLU modules perform downstream tasks like intent recognition and slot filling from the transcripts obtained. Creating language specific automatic speech recognition (ASR) modules for each language requires a large amount labelled data, which is usually not available for most languages. Language specific ASR systems thus form a bottleneck for creating SLU systems for low-resourced languages. %SLU are thus highly dependent on text transcriptions which form a bottleneck in creating such systems for low-resourced languages. 

In this paper, we present a novel acoustics based SLU system where we bypass the need to create a language specific ASR system. A block diagram representing our proposed system is shown in Figure \ref{fig:NOW}. We replace language specific STT modules with Allosaurus \cite{li2020universal}, which is a universal phone recognition system that creates phonetic transcription of input speech. We then create language and task specific, word-free, natural language understanding modules that perform NLU tasks like intent recognition and slot filling from phonetic transcriptions. 

%Re write this. You've added new languages
In this paper, we show that our proposed SLU system performs competitively with current state of the art SLU systems in high resource setting, and gets better as the amount of data available to the system reduces. We train both recurrent and transformer based neural networks and compare their performance for the task of intent classification. We work with natural speech datasets in five languages - English, Belgian Dutch, Mandarin, Sinhala and Tamil. Our system improves on the state of the art intent classification accuracy by approximately 5\% for Sinhala and 11\% for Tamil in low resource settings. We also propose an unsupervised slot value identification algorithm based on the self-attention mechanism. This enables one-shot data generation for a new slot value, where only one speech utterance is needed to generate new data.

\section{Related Work}
Spoken language understanding (SLU) systems are a vital component of spoken dialog systems. These systems are responsible for understanding the meaning of a spoken utterance. Doing so requires identifying speaker intent, a task which sometimes requires slot filling. Current research in spoken language understanding is moving towards creating End-to-End (E2E) SLU systems \cite{qian2017exploring} \cite{serdyuk2018towards} \cite{chen2018spoken} which have various advantages over conventional SLU systems \cite{lugosch2019speech}. To aid development in SLU, various speech to user intent datasets have been created in different languages including English \cite{lugosch2019speech} \cite{wu2020harpervalleybank} \cite{hemphill1990atis} \cite{saade2018spoken}, Sinhala \cite{buddhika2018domain} \cite{karunanayake2019transfer}, Tamil \cite{karunanayake2019transfer}, Belgian Dutch \cite{renkens2014acquisition} \cite{renkens2018capsule}, Mandarin \cite{zhu2019catslu} and French \cite{saade2018spoken}. For our work, we choose English, Sinhala, Tamil, Belgian Dutch and Mandarin datasets.

In low-resourced scenarios, building language specific ASR systems is not viable. In previous work, NLU modules have been built on top of outputs of an English ASR system, for example, using the softmax outputs of DeepSpeech \cite{hannun2014deep} for Sinhala and Tamil. DeepSpeech is a character level model and the softmax outputs corresponding to the model vocabulary were used as inputs to the intent classification model \cite{karunanayake2019transfer}. Softmax outputs of an English phoneme recognition system \cite{lugosch2019speech} have also been used to build intent recognition systems \cite{karunanayake2019sinhala} for Sinhala and Tamil. MFCC features of input speech have also been used for intent classification in Sinhala \cite{buddhika2018domain}. 

In our work, we build a unique natural language understanding module for SLU systems based on phonetic transcriptions of audio. These phonetic transcriptions were obtained from Allosaurus \cite{li2020universal}, a universal phone recognition system that gives language and speaker independent phonetic transcriptions of input audio. These transcriptions are finer grained when compared to a language specific phonemic transcription. This can be seen in the experiments sections where using only the top-1 prediction made by Allosaurus improves the performance on Sinhala and Tamil, which previously used the entire softmax vector of an English ASR system. The advantage of using Allosaurus to generate phonetic transcriptions are manifold. Allosaurus is trained to perform universal phone recognition, and is not a language specific model. This means the phonetic transcriptions encode finer grained information when compared to English phonemic representations. Also, these phonetic transcriptions incorporate language specific nuances and is expected to generalize better to novel languages, especially to languages that are phonetically different from English. 

A prototypical intent classification system was built for banking domain in Hindi from these phonetic transcriptions \cite{gupta2020mere}. In this work, a small natural speech dataset was used with Naive Bayes classifier as the intent recognition model. \cite{gupta2020acoustics} showed that such intent recognition systems built on top of phonetic transcriptions work for a large number of languages, including various Indic and Romance languages. They also showed that multilingual training helps in building more robust systems and improves performance on an unknown language within the same language family. The intent classification system described in \cite{gupta2020acoustics} was built for a large dataset with multiple intents, but this system was built using synthetic speech. In this paper, we perform intent classification and slot identification experiments on standard SLU datasets with natural speech. These are the first results for our proposed SLU system on natural speech.

%Write a paragraph on transformer models. Introduce BERT and RoBERTa and then write that your model training was based on roberta.
Transformer \cite{vaswani2017attention} based architectures have achieved state of the art performance in various speech and natural language processing tasks. BERT \cite{devlin2018bert} is a transformer based contextualized word embedding model which pushed the boundaries on performance on various NLP tasks including classification, natural language inference and question answering. BERT consists of the encoder modules of the Transformer, trained on the Masked Language Modelling (MLM) and the Next Sentence Prediction (NSP) objectives. RoBERTa \cite{liu2019roberta}, makes various modifications to the original BERT model including removing the NSP objective. In our paper, we train a RoBERTa based model with a vocabulary of phones.

\begin{table*}
\centering
\begin{tabular}{c|c|c|c|c}
%\hline
\multicolumn{1}{p{2cm}|}{\centering \textbf{Language}} &  \multicolumn{1}{|p{3cm}|}{\centering \textbf{Number of Utterances}} &  \multicolumn{1}{|p{3cm}|}{\centering \textbf{Number of Intents}} & \multicolumn{1}{|p{3cm}}{\centering \textbf{Number of Speakers}} & \multicolumn{1}{|p{3cm}}{\centering \textbf{Pre-Training Dataset Size (hrs)}}\\
\hline
English \cite{lugosch2019speech} & 30,043 & 31 &  97 & 100 \cite{panayotov2015librispeech} \\
Sinhala \cite{karunanayake2019transfer} & 7624 & 6 &  215 & 17 \cite{kjartansson2018crowd}\\
Tamil \cite{karunanayake2019transfer} & 400 & 6 &  40 & 7\cite{he2020open}\\
Belgian Dutch \cite{renkens2014acquisition} & 5940 & 36 & 11 & 63 \cite{kohn2016mining}\\
Mandarin \cite{zhu2019catslu} & 6925 & 4 & - & 15 \cite{wang2015thchs}
%\hline
\end{tabular}
\caption{Dataset statistics for English, Sinhala and Tamil datasets.}\label{Table1}
\end{table*}

\section{Datasets} \label{sec:datasets}
%%%%%%%%Write a sample intent from each dataset. Also write about the dutch and the mandarin dataset. Especially about the mandarin dataset and how you created it. 
%%%%%%Also write about pre-training datasets and a table for that

%write about datasets, how they were created. crowdsourcing etc.
We work with five standard SLU datasets for five different languages. All of these are natural speech datasets. For English, we use the Fluent Speech Commands (FSC) dataset \cite{lugosch2019speech}, which is the largest freely available speech to intent dataset. The dataset was collected using crowdsourcing and was also validated by a separate set of people by crowdsourcing. The dataset has 248 distinct sentences spoken by 97 different speakers. The FSC dataset was further divided into train, validation and test splits by the respective authors, where the validation and test sets comprised exclusively of 10 speakers which were not included in the other splits \cite{lugosch2019speech}. Detailed dataset statistics are shown in Table \ref{Table1}. Each utterance in the FSC dataset has three types of slot values for \textit{action}, \textit{object} and \textit{location}. The dataset can be modelled as a multilabel classification problem or a standard classification problem that flattens out all the different intents and slot values \cite{lugosch2019speech} \cite{radfar2020end}. We have used the 31-class intent classification formulation of the problem in our work. An example utterance in the dataset is given below:
\begin{itemize}
  \item[] Utterance: \textit{Switch the lights on in the kitchen} 
  \item[](action: activate), (object: lights), (location: kitchen)
\end{itemize}

The Belgian Dutch dataset, Grabo, \cite{renkens2014acquisition} was collected by asking users to control a service robot. There are 36 commands spoken by 11 different speakers which typically look like ``move to position $x$" or ``grab object $y$". The dataset was divided into a ratio of 60-20-20 for training, validation and testing. For detailed dataset statistics, please refer to Table \ref{Table1}. Here, each speaker was asked to say the same utterance 15 times. Thus, each intent class contains the exact same utterance repeated many times.

For Mandarin, we modify the CATSLU dataset \cite{zhu2019catslu} to make it suitable for intent classification. The original CATSLU dataset is ideal for dialog state tracking with conversations about 4 domains - Navigation, Music, Video and Weather. We convert the conversations into a 4-class intent classification dataset into the above four domains. To do this, we chose utterances corresponding to the semantic labels of the above domains as labels. The dataset statistics are shown in Table \ref{Table1}. This dataset contains longer and free-flowing sentences when compared to the other datasets. Examples of utterances for the class of \textit{weather} are:
\begin{itemize}
  \item[] \textit{what’s the weather in Shanghai today}, (Intent: Weather) 
  \item[] \textit{Is it sunny tomorrow}, (Intent: Weather)
\end{itemize}

We see that the above utterances are much more complex than utterances in other datasets and requires inferring that \textit{`Is it sunny tomorrow?'} corresponds to the domain of weather even though the word `weather' is not present in the utterance. This makes this dataset the most complex out of all the datasets used in our work. This is also shown by the fact that a BERT-based textual intent classification model achieves a classification accuracy of only 93\% and F1 score of 91\%.

We also work with speech to intent datasets in Sinhala \cite{buddhika2018domain} \cite{karunanayake2019transfer} and Tamil \cite{karunanayake2019transfer}. The Sinhala and Tamil datasets contain user utterances for a banking domain. The dataset has 6 different intents to perform common banking tasks including money withdrawal, deposit, credit card payments etc. Both datasets were collected via crowdsourcing. The Sinhala and Tamil datasets were not divided into train and test splits by the respective authors and previous work and results provided in literature on these datasets are based on 5-fold cross-validation \cite{buddhika2018domain} \cite{karunanayake2019sinhala} \cite{karunanayake2019transfer}. The detailed dataset statistics are also shown in Table \ref{Table1}. The utterances in the Tamil dataset are at times code-mixed with English.

%Write about why Mandarin is the hardest of the three datasets

We also pre-train our models using large speech corpus released for public use. The hours of data used for pre-training is shown in Table \ref{Table1}. We pass the speech utterances present in pre-training corpuses through Allosaurus to obtain their phonetic transcriptions. These transcriptions are used to pre-train our models.

\section{Models} \label{sec:Model}
%%Make three sections - Transformer models, LSTM, Self-Attention
%% Expand on model. Write about transformer models and pre training.

To the best of our knowledge, in this paper we train the first BERT-based language models for a vocabulary of phones. The transformer model we use is based on RoBERTa \cite{liu2019roberta}. We use the CLS token for generating sentence level representation for the input utterance, which is used for classification. We do a grid search for hyper parameters like number of attention heads, hidden layer size for the feed forward layers and number of encoder layers. We refer the reader to the RoBERTa \cite{liu2019roberta} and Transformer \cite{vaswani2017attention} papers for architectural details. For pre-training the transformer, we use the MLM objective where 15\% of the tokens are randomly masked. Out of those, 80\% tokens and randomly changed to the token MASK, 10\% tokens are changed to a random token and the remaining 10\% are kept the same. 

\begin{figure}[t]
  \centering
  \includegraphics[width=0.7\linewidth]{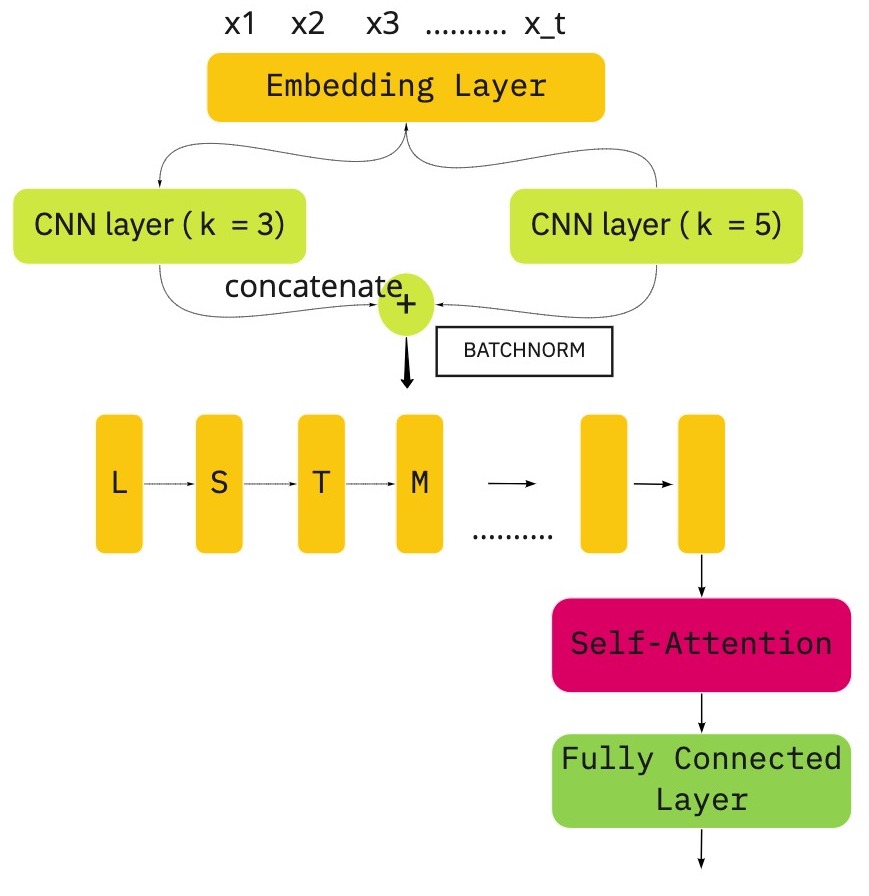}
  \caption{Model used for unsuperised slot identification.}
  \label{fig:Model}
\end{figure}

%previous work used softmax outputs. Our work is learning our own embeddings
We modify the language model architecture proposed in \cite{gupta2020acoustics} for our work. The architecture proposed in \cite{gupta2020acoustics} consists of a Convolutional Neural Network (CNN) and Long-Short Term Memory (LSTM) \cite{hochreiter1997long} based language model followed by a fully connected classification layer attached to the final time step output of the LSTM. The language model consists of CNN layers with varying filter sizes, capturing N-gram like features of word embeddings, very similar to the architecture shown in Figure \ref{fig:Model}. The CNN layers are followed by an LSTM layer. The CNN+LSTM layer forms the language model of our phonetic transcriptions. For intent classification, the LSTM output at the final time step is fed into a fully connected layer to perform intent classification. 

In addition to intent classification, we also propose an algorithm for unsupervised slot value identification, leveraging the self-attention mechanism \cite{bahdanau2014neural} to do so. We use the standard key-query-value formulation of the self-attention mechanism. The keys and values are the outputs of the LSTM at each time step, and the final time step output of the LSTM is used as the query. We use dot-product attention between query and key to calculate the attention scores. A softmax is taken across the attention scores for normalization. The final output of the self-attention mechanism is a linear combination of \textit{values} weighted by their normalized attention scores. The output of the self-attention layer is sent to the fully connected layer for the classification decision, as shown in Figure \ref{fig:Model}.

The phonetic transcriptions of all intent classification datasets which were used as training data including the language specific data splits will be made available publicly along with the codes used in this paper. The specific architectures used for achieving the results in section \ref{sec:experiments} will also be released along with the codes.

\begin{table*}
\centering
\begin{tabular}{c|c|c|c|c|c}
%\hline
\multicolumn{1}{p{1.5cm}|}{\centering \textbf{Dataset}} &  \multicolumn{1}{|p{1.5cm}|}{\centering \textbf{Baseline Accuracy}} &  \multicolumn{1}{|p{2cm}}{\centering \textbf{LSTM Model}} & \multicolumn{1}{|p{2cm}}{\centering \textbf{LSTM + Pre-Training}} & \multicolumn{1}{|p{2cm}}{\centering \textbf{Tranformer Model}} & \multicolumn{1}{|p{3cm}}{\centering \textbf{Tranformer + Pre-Training}}\\
\hline
English (FSC)  & 98.8\% \cite{lugosch2019speech} & 92.67 \% & 92.77 \% & 90.77 \% & 90.91 \% \\
Belgian Dutch (Grabo) & 94.5 \% \cite{renkens2018capsule} & 78.82 \% & 79.69 \% & 85.41 \% & 87.84 \% \\
Mandarin & - & 65.59 \% & 70.35 \% & 64.29 \% & 65.14 \% \\
Sinhala  & 97.31\% \cite{karunanayake2019sinhala} & 95.68 \% & 96.33 \% & 95.60 \% & 94.66 \% \\
Tamil & 81.7\% \cite{karunanayake2019sinhala} & 92.00 \% & 91.50 \% & 91.00 \% & 92.50 \%\\
%\hline
\end{tabular}
\caption{Intent classification results for our proposed SLU system.}\label{Table3}
\end{table*}

\section{Experiments}\label{sec:experiments}
%Note that we're just using the top-1 prediction.
%we should do real low resource experiment, maybe k shot or 10 % experiments, to show it works in low resource scenarios

\subsection{Intent Classification}
\label{sec:intents}
%All results
%Results with top 5
%Results with real k-shot experiments
%write about loss functions, optimizer, learning rate etc
%show the fact that English results are E2E results and thus are expected to be more robust
%Make two sections, one for language model pre training, and the other for using top 5 
%Give a reason as to why its better in tamil - maybe when amount of data increases, the model is able to learn finer grained differences from the data itself, which are provided by Allosaurus
We compare the performance of our proposed system with previously reported results on five different languages - English, Belgian Dutch, Mandarin, Sinhala and Tamil. The results are shown in Table \ref{Table3}. We see that the results of our proposed method improves as the size of the dataset decreases. This makes our system an ideal candidate to be used for low-resourced scenarios. We report state of the art results for Tamil, which improves on the previous best by approximately 11\% and halves the error rate. %All baseline results in their original papers have been reported with the accuracy metrics. To compare with previous work, our results are also presented using the accuracy metric. 

%The results are competitive for the Fluent Speech Commands dataset. Our results are very close to the previously best reported results for the Sinhala dataset \cite{karunanayake2019sinhala}. Our system outperforms previously reported results on the Tamil dataset by a significant margin. We improve on previous reported accuracy by more than 10\% and half the error rate. All baseline results in their original papers have been reported with the accuracy metrics. To compare with previous work, our results are also presented using the accuracy metric. 

%The FSC dataset was divided into a train, validation and test set by the respective authors \cite{lugosch2019speech}. The test set contains the phonetic transcriptions of 10 different speakers which were not present in the train or the validation set. The results shown in Table \ref{Table3} are for the test set. The speaker information was not present in the Sinhala and Tamil dataset. The results reported in previous works \cite{karunanayake2019sinhala} \cite{karunanayake2019transfer} are average results for 5-fold cross-validation. We employ the same strategy in our experiments. 

Our proposed phonetic transcription based intent classifier performs competitively on the relatively larger English dataset. The model trained on the FSC dataset in \cite{lugosch2019speech} is an E2E-SLU model, which has various advantages over a two-module split SLU system. An E2E-SLU model learns better representation of data as it directly optimizes for the metric of intent classification \cite{lugosch2019speech}. Instead, our system consists of two blocks that are not optimized for the errors made by the other, causing errors to propagate through the system. Thus, an end-to-end system with enough data provides an upper limit for the intent classification results. Furthermore, results shown in Table \ref{Table3} only use the top-1 predictions made by Allosaurus, which means we select the phone with the highest softmax score for generating phonetic transcriptions. When we use the top-5 predictions made by Allosaurus, thus giving more information about the spoken utterance to our models, we achieve an accuracy of 96.31\%. 

%Dutch - write why dutch show such high improvements with transformers
The Belgian Dutch dataset observes significant improvement with the use of Transformer models when compared to the other datasets. We attribute this improvement to the kinds of utterance present in the dataset. All utterances corresponding to the same intent in the Dutch dataset are spoken in the exact same word order. The positional embeddings of a transformer are responsible for encoding the grammatical structure of an utterance. When the structure is jumbled, as in the case of the other datasets which contain multiple ways of saying the same intent, the positional embeddings don't encode useful information and require larger amounts of data. Also, note that since multiple phones can correspond to the same spoken word, the token order in phonetic transcriptions are even more jumbled. Since the Dutch dataset does not have jumbled word order, there is minimal variability in the dataset and hence transformers are able to produce better results for Belgian Dutch when compared to other languages. 

%About Mandarin, write about why the performance is so low. Explain why pre-training might have helped here.
The Mandarin dataset is by far the toughest dataset. The other four datasets have simple commands with most sentence lengths of 2-5 words. The Mandarin dataset on the other hand contains free-flowing questions which are also longer in length. The utterances do not contain intent specific words as shown in section \ref{sec:datasets} which makes the task harder. This is why we see modest classification performance for the Mandarin dataset.

The baseline models for the Sinhala and Tamil datasets \cite{karunanayake2019sinhala} are a two-module split SLU systems, where the intent classifier is built from phonemic transcriptions generated from an English ASR. The performance of our system is comparable for the Sinhala dataset while it significantly outperforms the phonemic transcription based model for the Tamil dataset. We again point out to the reader that in our systems, we've only used the top-1 softmax predictions made by Allosaurus, thus providing much less information about spoken utterances to our model. On the other hand, baseline models use the entire softmax layer vectors of the ASR systems as an embedding to encode spoken information. 

%crosslingually
The above results also show the effectiveness of using Allosaurus when compared to a language specific ASR for encoding spoken information. Phonemes are perceptual units of sounds and changing phonemes ends up changing the spoken word. Phones on the other hand are language independent and correspond to the actual sound produced. Changing a phone does not necessarily change a word in a particular language. Usually multiple phones are mapped to a single phoneme, and this mapping is language specific. This makes using ASR systems built for a high-resourced language like English sub-optimal when cross-lingually encoding spoken utterances into a vector for chosen target languages. Allosaurus \cite{li2020universal} is a universal phone recognition model and is trained to recognize fine grained differences in spoken utterances at the level of phones. We hypothesize that the intent classification models are able to extract fine-grained phone level differences when larger amounts of data is available, but as the amount of data reduces, this becomes increasingly difficult. This is why we see improvement in performance as the dataset size decreases.

\begin{figure}[t]
  \centering
  \includegraphics[width=0.7\linewidth]{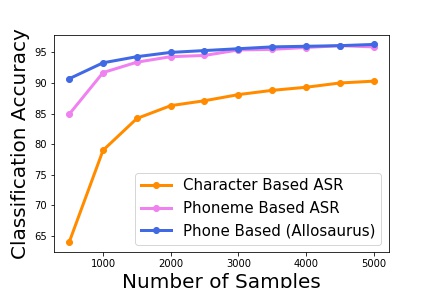}
  \caption{Comparing the performance of our proposed phonetic transcription based SLU system with previous characted and phone based systems.  }
  \label{fig:comparison}
\end{figure}

%show figure for dataset difference experiments from 500 to 5000.
To illustrate the effectiveness of our proposed system over previous approaches, we randomly select subsets of the utterances in Sinhala in increments of 500, just as it was done in \cite{karunanayake2019transfer} \cite{karunanayake2019sinhala}. We compare the performance of our phonetic transcription based model to the English character based ASR model \cite{karunanayake2019transfer} and the English phoneme based ASR model \cite{karunanayake2019sinhala}. The authors very generously provided us with the exact numbers for the comparison. The results can be seen in Figure \ref{fig:comparison} . We can see that intent classifiers built from phonetic transcriptions produced by Allosaurus significantly outperforms intent recognition systems built on top of character or phoneme based transcriptions as the amount of data reduces. For a training set size of 500 samples for the Sinhala dataset, we improve on the English character based system by approximately 25\% and on the English phoneme based system by more than 5\%. The transformer and pre-trained models do not outperform the CNN+LSTM models, especially for low data scenarios. 

%%Write final word on transformer and pre-training

\subsection{Unsupervised Slot Value Identification}
%emphasize that this we use attention to do the unsupervised slot id
%application, data augmentation using a single sample
%creating a new slot/class if slot is identified
%write an intro of what you're doing once you have a sense of it
%Results in section \ref{sec:intents} clearly show that our proposed intent recognition system based on phonetic transcriptions works for high resourced scenarios and shows superior performance as the amount of available labelled data decreases. 

With the aim of creating entire NLU modules based on phonetic transcriptions of speech, we shift our focus on slots. The Sinhala and Tamil datasets used in this paper do not have sequence level slot information either in speech or textual transcriptions. This is a very realistic low-resource scenario where we cannot expect utterances to have labelled slot values. Similar argument holds for unwritten languages. In this section, we propose an attention based \textbf{L}ow-Resource \textbf{U}nsupervised \textbf{S}lot value \textbf{Id}entification (LUSID) algorithm to identify slots values when no labels are present. We also aim to identify the span of existing slots in our training data (note that our training data is phonetic transcriptions of speech).

%Using this, we can create artificial training data in phonetic transcriptions for a new slot values by replacing it in the found slot locations. To do so, we would require only one speech sample of the new slot value, whose phonetic transcription we can generate through Allosaurus. %This method can also be used for augmenting existing data by creating artificial data in the phonetic transcription domain without the need for additional speech recording.  

%with the aim of being able to replace them to generate artificial data for a new slot unseen by our model. This method can also be used for augmenting existing data by creating artificial data in the phonetic transcription domain without the need for additional speech recording, thus boosting any dataset with speech recordings. The artificial data is referred to as the \textit{boosted data}. 

%We model the slot identification problem as a classification problem. The aim here is to identify span of the slots in the training data with the aim of being able to replace them to generate artificial data for a new slot. This method can also be used for augmenting existing data by creating artificial data in the phonetic transcription domain without the need for additional speech recording, thus boosting any dataset with speech recordings. The artificial data is referred to as the \textit{boosted data}. 

\subsubsection{Problem Definition}
We pose the unsupervised slot value identification as a classification problem. We use an attention based classification model as described in section \ref{sec:Model} to identify slot values in an unsupervised manner. A self-attention module is added before the final classification layer of the LSTM+CNN based model. To test our algorithm, we create a 2-class attention-based classification model. The two classes correspond to two slot values belonging to the same intent, thus the differentiating feature is the slot value between them. 

We use an example from the FSC dataset for illustration. The intent of \textit{activating lights} is used from the FSC dataset with two slot values - \textit{bedroom} and \textit{kitchen}. Figure \ref{fig:Slot} shows the normalized attention scores for a given phonetic input when the utterances are passed through the attention-based classifier. The title of the figure represents the textual transcription of the speech input. The x-axis labels correspond to the phonetic transcription of the input produced by Allosaurus. The y-axis represent the normalized attention scores for each token in the phonetic transcription. Figure \ref{fig:Slot} shows activated weights for phones corresponding to the word bedroom. We can see that the self-attention mechanism is able to identify the approximate location of the slot value for the slot \textit{location}. 

\subsubsection{Identifying Exact Slot Location}
Next, we identify the location of slot values in an utterance. This would enable us to replace the slot value with a new slot value, thus generating synthetic data. For the purpose of this illustration, let's say our base slot value is \textit{bedroom} and the target slot value is \textit{kitchen}. \textit{Base slot value} is the slot value in the utterance we're working with. This is the slot value we want to replace in the current utterance with the \textit{target slot value}, keeping the remaining utterance the same. To do this, we identify the phone corresponding to the highest attention score for the base slot value, and remove all phones within a left window of size $l=4$ phones and a right window of size $r=3$ phones from the highest score phone. This gives us the location of the base slot value. $(l,r)$ are tuned as hyperparameters for each model. Once we have the location of the base slot value, we replace it with the target slot value (corresponding to the phonetic transcription of \textit{kitchen}). 

\subsubsection{Verification of Generated Data}
The above process gives us a synthetically generated utterance for the slot value \textit{kitchen} from an utterance corresponding to the slot value of \textit{bedroom}. Note that this new utterance is generated purely in the phonetic transcription domain, thus avoiding the need for textual transcriptions and supervised slot level labelling. Next, we need to test that this new generated utterance actually corresponds to the target slot value, \textit{kitchen}. To do so, we feed the generated utterance back to the same classifier, with the expectation of now being classified into the target slot value class, \textit{kitchen}. The accuracy for the new utterance generated from the base slot value, being classified as the target slot value, is shown in Table \ref{Table2} with optimal $(l,r)$ values for both the English and the Sinhala dataset. For the English slots, the best classification accuracy achieved is 99.24 \%, which means that the model classifies the generated data into the target class 99.24\% time. This shows that the model is not able to differentiate between generated utterances and actual data. For the Sinhala dataset, we used the intents \textit{bill Payments} and \textit{credit card payments} and achieved an accuracy of 93.61 \%. 

\subsubsection{Discussion}
\begin{figure}[t]
  \centering
  \includegraphics[width=\linewidth]{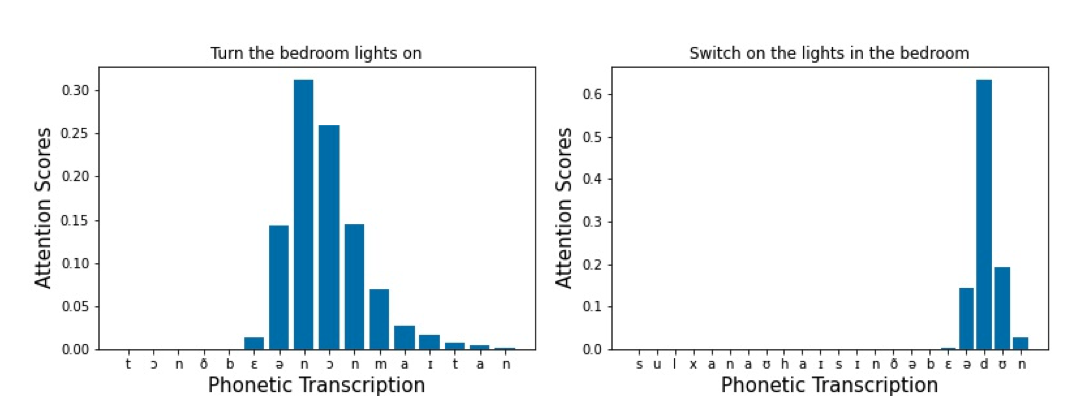}
  \caption{Attention scores for each phone in the phonetic transcription of an utterance.}
  \label{fig:Slot}
\end{figure}

It is important to highlight the non-triviality of these results. Firstly, the slot can be present anywhere in the sentence and we identify the span of the slot in an unsupervised way. This can be seen in Figure \ref{fig:Slot}. Secondly, taking the example of the English dataset, the generated utterances are created from the training data for base slot value of the \textit{bedroom} class. Thus, the model is trained with almost 100\% accuracy to recognize the non-replaced part of the utterance as the \textit{bedroom} class. Yet, it recognizes new utterance generated from the base class (\textit{bedroom}) as belonging to the target class (\textit{kitchen}), showing that we have successfully removed the slot value which was responsible for making the classification decision.

LUSID can be used to generate artificial data for a new slot value from a single spoken utterance. We can also use this algorithm to generate unsupervised slot labels in the phonetic transcription domain when slot labels are not present as well as for data augmentation, since it allows us to generate new data samples for a given slot value for an existing dataset. 

\begin{table}
\centering
\begin{tabular}{c|c|c|c}
%\hline
\multicolumn{1}{p{1.5cm}|}{\centering \textbf{Dataset}} &
\multicolumn{1}{p{1.5cm}|}{\centering \textbf{Left \\Window}} &  \multicolumn{1}{|p{1.5cm}|}{\centering \textbf{Right\\ Window}} &  \multicolumn{1}{|p{2cm}}{\centering \textbf{Classification Accuracy}}\\
\hline
English (FSC) & 4 & 3 & 99.24 \%\\
Sinhala & 8 & 1 & 93.61 \%\\
%\hline
\end{tabular}
\caption{Classification accuracy of generated utterances using LUSID. }\label{Table2}
\end{table}

\section{Conclusions}
In this paper, we present a unique spoken language understanding system (SLU) for low-resourced and unwritten languages. The SLU system converts speech to its phonetic transcription using a universal speech to phone converter. We then build natural language understanding modules for utterances in the phonetic transcriptions domain, which perform competitively with current end-to-end SLU models and outperforms state of the art approaches for low-resourced languages. Moreover, we show that the performance of our system surpasses state of the art systems as the amount of labelled data decreases, which makes it an ideal candidate for low-resourced settings. We also propose an attention-based unsupervised slot value identification algorithm that identifies slots in the phonetic transcription domains when slot labels are not present. This technique, which we call LUSID, has various applications from one-shot data generation to data augmentation. 
%Our method has the potential to work for unwritten languages, focus on no words
%Do not need a speech to text system
%dont even need text
%work with other languages which will pose their own challeneges (write about phonetic challenges like chinese)
%next main thing is to deal with slot filling
%creating a fully functioning SLU system that is competitice for high resourced languages and improves performance in low resource languages
%Allosaurus is easy to use

\bibliographystyle{IEEEbib}
\bibliography{strings,refs}

\begin{thebibliography}{10}

\bibitem{li2020universal}
Xinjian Li, Siddharth Dalmia, Juncheng Li, Matthew Lee, Patrick Littell, Jiali
  Yao, Antonios Anastasopoulos, David~R Mortensen, Graham Neubig, Alan~W Black,
  et~al.,
\newblock ``Universal phone recognition with a multilingual allophone system,''
\newblock in {\em ICASSP 2020-2020 IEEE International Conference on Acoustics,
  Speech and Signal Processing (ICASSP)}. IEEE, 2020, pp. 8249--8253.

\bibitem{qian2017exploring}
Yao Qian, Rutuja Ubale, Vikram Ramanaryanan, Patrick Lange, David
  Suendermann-Oeft, Keelan Evanini, and Eugene Tsuprun,
\newblock ``Exploring asr-free end-to-end modeling to improve spoken language
  understanding in a cloud-based dialog system,''
\newblock in {\em 2017 IEEE Automatic Speech Recognition and Understanding
  Workshop (ASRU)}. IEEE, 2017, pp. 569--576.

\bibitem{serdyuk2018towards}
Dmitriy Serdyuk, Yongqiang Wang, Christian Fuegen, Anuj Kumar, Baiyang Liu, and
  Yoshua Bengio,
\newblock ``Towards end-to-end spoken language understanding,''
\newblock in {\em 2018 IEEE International Conference on Acoustics, Speech and
  Signal Processing (ICASSP)}. IEEE, 2018, pp. 5754--5758.

\bibitem{chen2018spoken}
Yuan-Ping Chen, Ryan Price, and Srinivas Bangalore,
\newblock ``Spoken language understanding without speech recognition,''
\newblock in {\em 2018 IEEE International Conference on Acoustics, Speech and
  Signal Processing (ICASSP)}. IEEE, 2018, pp. 6189--6193.

\bibitem{lugosch2019speech}
Loren Lugosch, Mirco Ravanelli, Patrick Ignoto, Vikrant~Singh Tomar, and Yoshua
  Bengio,
\newblock ``Speech model pre-training for end-to-end spoken language
  understanding,''
\newblock {\em arXiv preprint arXiv:1904.03670}, 2019.

\bibitem{wu2020harpervalleybank}
Mike Wu, Jonathan Nafziger, Anthony Scodary, and Andrew Maas,
\newblock ``Harpervalleybank: A domain-specific spoken dialog corpus,''
\newblock {\em arXiv preprint arXiv:2010.13929}, 2020.

\bibitem{hemphill1990atis}
Charles~T Hemphill, John~J Godfrey, and George~R Doddington,
\newblock ``The atis spoken language systems pilot corpus,''
\newblock in {\em Speech and Natural Language: Proceedings of a Workshop Held
  at Hidden Valley, Pennsylvania, June 24-27, 1990}, 1990.

\bibitem{saade2018spoken}
Alaa Saade, Alice Coucke, Alexandre Caulier, Joseph Dureau, Adrien Ball,
  Th{\'e}odore Bluche, David Leroy, Cl{\'e}ment Doumouro, Thibault
  Gisselbrecht, Francesco Caltagirone, et~al.,
\newblock ``Spoken language understanding on the edge,''
\newblock {\em arXiv preprint arXiv:1810.12735}, 2018.

\bibitem{buddhika2018domain}
Darshana Buddhika, Ranula Liyadipita, Sudeepa Nadeeshan, Hasini Witharana,
  Sanath Javasena, and Uthayasanker Thayasivam,
\newblock ``Domain specific intent classification of sinhala speech data,''
\newblock in {\em 2018 International Conference on Asian Language Processing
  (IALP)}. IEEE, 2018, pp. 197--202.

\bibitem{karunanayake2019transfer}
Yohan Karunanayake, Uthayasanker Thayasivam, and Surangika Ranathunga,
\newblock ``Transfer learning based free-form speech command classification for
  low-resource languages,''
\newblock in {\em Proceedings of the 57th Annual Meeting of the Association for
  Computational Linguistics: Student Research Workshop}, 2019, pp. 288--294.

\bibitem{renkens2014acquisition}
Vincent Renkens, Steven Janssens, Bart Ons, Jort~F Gemmeke, et~al.,
\newblock ``Acquisition of ordinal words using weakly supervised nmf,''
\newblock in {\em 2014 IEEE Spoken Language Technology Workshop (SLT)}. IEEE,
  2014, pp. 30--35.

\bibitem{renkens2018capsule}
Vincent Renkens et~al.,
\newblock ``Capsule networks for low resource spoken language understanding,''
\newblock {\em arXiv preprint arXiv:1805.02922}, 2018.

\bibitem{zhu2019catslu}
Su~Zhu, Zijian Zhao, Tiejun Zhao, Chengqing Zong, and Kai Yu,
\newblock ``Catslu: The 1st chinese audio-textual spoken language understanding
  challenge,''
\newblock in {\em 2019 International Conference on Multimodal Interaction},
  2019, pp. 521--525.

\bibitem{hannun2014deep}
Awni Hannun, Carl Case, Jared Casper, Bryan Catanzaro, Greg Diamos, Erich
  Elsen, Ryan Prenger, Sanjeev Satheesh, Shubho Sengupta, Adam Coates, et~al.,
\newblock ``Deep speech: Scaling up end-to-end speech recognition,''
\newblock {\em arXiv preprint arXiv:1412.5567}, 2014.

\bibitem{karunanayake2019sinhala}
Yohan Karunanayake, Uthayasanker Thayasivam, and Surangika Ranathunga,
\newblock ``Sinhala and tamil speech intent identification from english phoneme
  based asr,''
\newblock in {\em 2019 International Conference on Asian Language Processing
  (IALP)}. IEEE, 2019, pp. 234--239.

\bibitem{gupta2020mere}
Akshat Gupta, Sai~Krishna Rallabandi, and Alan~W Black,
\newblock ``Mere account mein kitna balance hai?--on building voice enabled
  banking services for multilingual communities,''
\newblock {\em arXiv preprint arXiv:2010.16411}, 2020.

\bibitem{gupta2020acoustics}
Akshat Gupta, Xinjian Li, Sai~Krishna Rallabandi, and Alan~W Black,
\newblock ``Acoustics based intent recognition using discovered phonetic units
  for low resource languages,''
\newblock {\em arXiv preprint arXiv:2011.03646}, 2020.

\bibitem{vaswani2017attention}
Ashish Vaswani, Noam Shazeer, Niki Parmar, Jakob Uszkoreit, Llion Jones,
  Aidan~N Gomez, {\L}ukasz Kaiser, and Illia Polosukhin,
\newblock ``Attention is all you need,''
\newblock in {\em Advances in neural information processing systems}, 2017, pp.
  5998--6008.

\bibitem{devlin2018bert}
Jacob Devlin, Ming-Wei Chang, Kenton Lee, and Kristina Toutanova,
\newblock ``Bert: Pre-training of deep bidirectional transformers for language
  understanding,''
\newblock {\em arXiv preprint arXiv:1810.04805}, 2018.

\bibitem{liu2019roberta}
Yinhan Liu, Myle Ott, Naman Goyal, Jingfei Du, Mandar Joshi, Danqi Chen, Omer
  Levy, Mike Lewis, Luke Zettlemoyer, and Veselin Stoyanov,
\newblock ``Roberta: A robustly optimized bert pretraining approach,''
\newblock {\em arXiv preprint arXiv:1907.11692}, 2019.

\bibitem{panayotov2015librispeech}
Vassil Panayotov, Guoguo Chen, Daniel Povey, and Sanjeev Khudanpur,
\newblock ``Librispeech: an asr corpus based on public domain audio books,''
\newblock in {\em 2015 IEEE international conference on acoustics, speech and
  signal processing (ICASSP)}. IEEE, 2015, pp. 5206--5210.

\bibitem{kjartansson2018crowd}
Oddur Kjartansson, Supheakmungkol Sarin, Knot Pipatsrisawat, Martin Jansche,
  and Linne Ha,
\newblock ``Crowd-sourced speech corpora for javanese, sundanese, sinhala,
  nepali, and bangladeshi bengali,''
\newblock 2018.

\bibitem{he2020open}
Fei He, Shan-Hui~Cathy Chu, Oddur Kjartansson, Clara Rivera, Anna Katanova,
  Alexander Gutkin, Isin Demirsahin, Cibu Johny, Martin Jansche, Supheakmungkol
  Sarin, et~al.,
\newblock ``Open-source multi-speaker speech corpora for building gujarati,
  kannada, malayalam, marathi, tamil and telugu speech synthesis systems,''
\newblock in {\em Proceedings of the 12th Language Resources and Evaluation
  Conference}, 2020, pp. 6494--6503.

\bibitem{kohn2016mining}
Arne K{\"o}hn, Florian Stegen, and Timo Baumann,
\newblock ``Mining the spoken wikipedia for speech data and beyond,''
\newblock in {\em Proceedings of the Tenth International Conference on Language
  Resources and Evaluation (LREC'16)}, 2016, pp. 4644--4647.

\bibitem{wang2015thchs}
Dong Wang and Xuewei Zhang,
\newblock ``Thchs-30: A free chinese speech corpus,''
\newblock {\em arXiv preprint arXiv:1512.01882}, 2015.

\bibitem{radfar2020end}
Martin Radfar, Athanasios Mouchtaris, and Siegfried Kunzmann,
\newblock ``End-to-end neural transformer based spoken language
  understanding,''
\newblock {\em arXiv preprint arXiv:2008.10984}, 2020.

\bibitem{hochreiter1997long}
Sepp Hochreiter and J{\"u}rgen Schmidhuber,
\newblock ``Long short-term memory,''
\newblock {\em Neural computation}, vol. 9, no. 8, pp. 1735--1780, 1997.

\bibitem{bahdanau2014neural}
Dzmitry Bahdanau, Kyunghyun Cho, and Yoshua Bengio,
\newblock ``Neural machine translation by jointly learning to align and
  translate,''
\newblock {\em arXiv preprint arXiv:1409.0473}, 2014.

\end{thebibliography}

\end{document}